\documentclass{article} 
\usepackage{iclr2018_conference,times}
\usepackage{hyperref}
\usepackage{url}
\usepackage{graphicx} 
\usepackage{amsmath}
\usepackage{amssymb}
\usepackage{subcaption}
\usepackage{color}
\usepackage[utf8]{inputenc} 
\usepackage[T1]{fontenc}    
\usepackage{booktabs}       
\iclrfinalcopy
\title{The (Un)reliability of saliency methods}
\newcommand{\bx}{\boldsymbol{x}}

\newcommand{\bs}{\boldsymbol{s}}
\newcommand{\bw}{\boldsymbol{w}}
\newcommand{\ba}{\boldsymbol{a}}
\newcommand{\by}{\boldsymbol{y}}
\newcommand{\bm}{\boldsymbol{m}}

\author{
Pieter-Jan Kindermans\footnotemark[2]{}, Sara Hooker\footnotemark[2]{}, Julius Adebayo \\
Google Brain\thanks{Work done as part of the Google Brain Residency program, Pieter-Jan Kindermans and Sara Hooker have contributed equally.}\\
  \texttt{\{pikinder, shooker\}@google.com} \\
    \AND
  Maximilian Alber, Kristof T. Sch\"{u}tt,  Sven D\"{a}hne\\
TU-Berlin\\
  \And
  Dumitru Erhan, Been Kim\\
Google Brain \\
}

\begin{document}

\maketitle

\begin{abstract}
 Saliency methods aim to explain the predictions of deep neural networks. These methods lack reliability when the explanation is sensitive to factors that do not contribute to the model prediction. We use a simple and common pre-processing step ---adding a constant shift to the input data--- to show that a transformation with no effect on the model can cause numerous methods to incorrectly attribute. In order to guarantee reliability, we posit that methods should fulfill input invariance, the requirement that a saliency method mirror the sensitivity of the model with respect to transformations of the input. We show, through several examples, that saliency methods that do not satisfy input invariance result in misleading attribution.
\end{abstract}

\section{Introduction}
While considerable research has focused on discerning the decision process of neural networks \citep{Baehrens2010,Simonyan2014,Haufe2014,Zeiler2014,Springenberg2014,Bach2015,Yosinski2015,Nguyen2016,Montavon2017,Zintgraf2017,Mukund2017,Smilkov2017,Kindermans2017}, there remains a trade-off between model complexity and interpretability. Research to address this tension is urgently needed; reliable explanations build trust with users, help identify points of model failure and remove barriers to entry for the deployment of deep neural networks in domains like health care, security and transportation.

In deep neural networks, data representation is delegated to the model and subsequently we cannot generally say in an informative way what led to a model prediction. Instead, saliency methods aim to infer insights about the $f(x)$ learnt by the model by ranking the explanatory power of constituent inputs. While unified in purpose, these methods are surprisingly divergent and non-overlapping in outcome. Evaluating the reliability of these methods is complicated by a lack of ground truth, as ground truth would depend upon full transparency into how a model arrives at a decision — the very problem we are trying to solve for in the first place.

Given the need for a quantitative method of comparison, several properties such as completeness, implementation invariance and sensitivity have been articulated as desirable to ensure that saliency methods are reliable \citep{Bach2015,Mukund2017}. Implementation invariance, proposed as an axiom for attribution methods by \citep{Mukund2017}, is the requirement that functionally equivalent networks (models with different architectures but equal outputs for all inputs), always attribute in an identical way.

This work posits that a second invariance axiom, which we term \textit{input invariance}, needs to be satisfied to ensure reliable interpretation of the input's contribution to the model prediction. \textit{Input invariance} requires that the saliency method mirror the sensitivity of the model with respect to transformations of the input. We demonstrate that numerous methods do not satisfy input invariance using a simple transformation -- a constant shift of the input -- that changes the attribution of these methods but does not affect the model prediction or weights. Our results demonstrate that explanations of a networks predictions can be purposefully manipulated using surprisingly simple transformations to be misleading. This work is motivated by an understanding that saliency methods are highly valued tools for gaining intuition about a network. Determining points of failure is a necessary step for knowledgeable use of these tools as well as a pre-requisite for domains like medicine where the incorrect classification of an input as salient carries a high cost.

\begin{figure}
\centering
    \includegraphics[width=\textwidth]{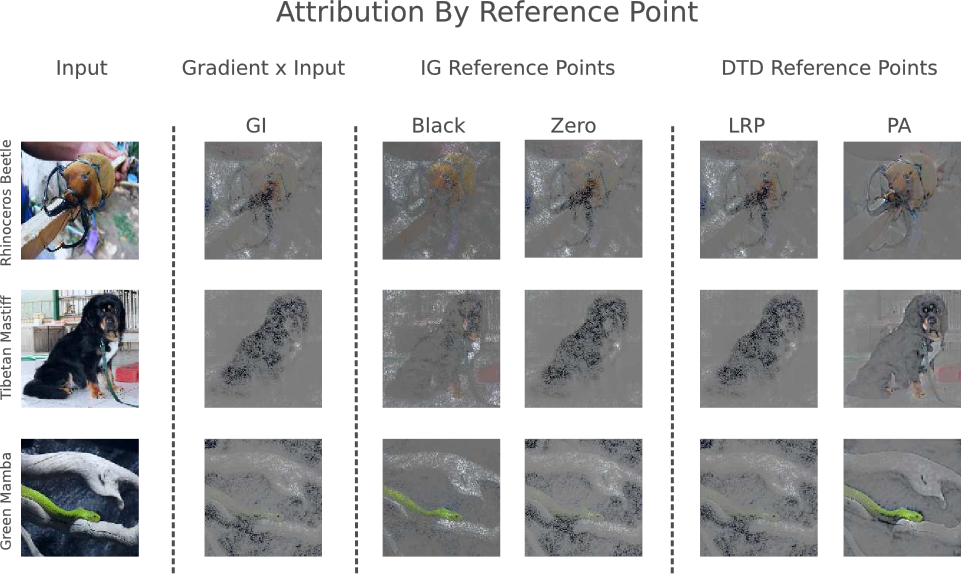}
\caption{
    Integrated gradients and Deep Taylor Decomposition determine input attribution relative to a chosen reference point. This choice determines the vantage point for all subsequent attribution. Using two example reference points for each method we demonstrate that changing the reference causes the attribution to diverge. The attributions are visualized in a consistent manner with the IG paper
    \protect\citep{Mukund2017}. Visualisations were made using ImageNet data. \protect\citep{Imagenet2015} and  the VGG16 architecture \protect\citep{Simonyan2014}.
    \label{fig:varyingrefpoint}}
\end{figure}

In this work we:
\begin{itemize}
\item introduce the axiom \textit{input invariance} and show, using a simple constant shift in the input, that certain saliency methods do not satisfy this property (See Fig.~\ref{fig:attributionmethods}).
\item demonstrate using MNIST that we can purposefully force misleading attribution (See Fig.~\ref{fig:square} and Fig.~\ref{fig:kitty}).
\item show that "reference point" methods -- Integrated gradients and the Deep Taylor Decomposition-- have diverging attribution satisfy input invariance contingent on the choice of reference and the type of transformation considered (See Fig.~\ref{fig:varyingrefpoint}).
\item propose data normalization as a way to ensure that some methods satisfy input invariance for the type of transformation considered. Discuss the need for wider research as normalization does not systematically guarantee reliable attribution for all possible transformations.
\end{itemize}

In \textbf{Section 2}, we detail our experiment framework. In \textbf{Section 3}, we determine that while the model is invariant to the input transformation considered, several saliency methods attribute to the mean shift. In \textbf{Section 4} we discuss "reference point" methods and illustrate the importance of choosing an appropriate reference before discussing some directions for future research in \textbf{Section 5}. 

\section{The model is invariant to a constant shift in input}\label{sec:shift}

We show that, by construction, the bias of a neural network compensates for the constant shift resulting in two networks with identical weights and predictions.

We compare the attribution across two networks, $f_1(x)$ and $f_2(x)$. $f_1(x)$ is a network trained on input $\bx^i_1$ that denotes sample $i$ from training set $X_1$. The classification task of network 1 is:
\[
f_1(\bx^i_1) = \by^i,
\]

$f_2(x)$ is a network that predicts the classification of a transformed input $\bx^i_2$. The relationship between $\bx^i_1$ and $\bx^i_2$ is the addition of constant vector $\bm_2$:
\begin{eqnarray*}
\forall i, \bx_2^i &=& \bx^i_1 + \bm_2.
\end{eqnarray*}

Network 1 and 2 differ only by construction. Consider the first layer neuron before non-linearity in $f_1(x)$:
\[
z = \boldsymbol{w}^T\bx_1+b_1. 
\]

We alter the biases in the first layer neuron by adding the mean shift $\bm_2$. This now becomes Network 2:
\[
b_2 = b_1 - \bw^T\bm_2.
\]

As a result the first layer activations are the same for $f_1(x)$ and $f_2(x)$: 
\[
z = \bw^T\bx_2+b_2=\bw^T\bx_1+\bw^T\bm_2+b_1-\bw^T\bm_2.
\]

Note that the gradient with respect to the input remains unchanged as well:
\[
 \frac{\partial f_1(\bx^i_1)}{\partial \bx^i_1} = \frac{\partial f_2(\bx^i_2)}{\partial \bx^i_2}.
\]

We have shown that Network 2 cancels out the mean shift transformation. This means that $f_1(x)$ and $f_2(x)$ have identical weights and produce the same output for all corresponding samples,  $x_1^i\in X_1$, $x_2^i\in X_2$: 
 
\[
\forall i, f_1(\bx^i_1)=f_2(\bx^i_2).
\]

\subsection{Experimental Setup}
Now, we describe our experiment setup to evaluate the input invariance of a set of saliency methods. Most saliency research to date has centered on convolutional neural networks (CNN). In this work, we also evaluate input invariance using a CNN. 
Network 1 is a 3 layer multi-layer perceptron with 1024 ReLu-activated neurons each. Network 1 classifies MNIST image inputs in a [0,1] encoding. We consider a negative constant shift of $\bm_2=-1$; Network 2 classifies MNIST image inputs in a [-1,0] MNIST encoding. The first network is trained for 10 epochs using mini-batch stochastic gradient descent (SGD). The final accuracy is 98.3\% for both\footnote[3]{Although there is a gap between this and the state of art, the gap does not significantly influence our findings.}. In \ref{subsection0} we introduce the saliency methods we evaluate. 

\section{The (In)sensitivity of Saliency Methods to Mean Shifts}

In \ref{subsection0} we introduce key approaches to the classification of inputs as salient and the saliency methods we evaluate. In \ref{subsection1} we find that gradient and signal methods satisfy input invariance. In \ref{subsection2} we find that all attribution methods considered have points of failure.

\subsection{Saliency methods considered} \label{subsection0}

Saliency methods broadly fall into three different categories:
\begin{enumerate}
 \item \textbf{Gradients (Sensitivity)} \citep{Baehrens2010,Simonyan2013} shows how a small change to the input affects the classification score for the output of interest.
 \item \textbf{Signal methods} such as DeConvNet \citep{Zeiler2014}, Guided BackProp \citep{Springenberg2014} and PatternNet \citep{Kindermans2017} aim to isolate input patterns that stimulate neuron activation in higher layers.
 \item \textbf{Attribution methods} such as Deep-Taylor Decomposition  \citep{Montavon2017} and Integrated Gradients \citep{Mukund2017} assign importance to input dimensions by decomposing the value $y_j$ at an output neuron $j$ into contributions from the individual input dimensions:
\[
\bs_j = A(\bx)_j.
\] $\bs_j$ is the decomposition into input contributions and has the same number of dimensions as $\bx$, $A(\bx)_j$ signifies the attribution method applied to output $j$ for sample $\bx$. Attribution methods are distinct from gradients because of the insistence on \emph{completeness}; the sum of all attributions should be approximately equal to the original output $y_i$.
\end{enumerate}

We consider the input invariance of each category separately (by evaluating raw gradients, GuidedBackprop, PatternNet, Integrated Gradients and Deep Taylor Decomposition) and also benchmark the input invariance of SmoothGrad \citep{Smilkov2017}, a method that wraps around an underlying saliency approach and uses the addition of noise to produce a sharper visualization of the saliency heatmap.

The experiment setup and methodology is as described in \textbf{Section 2}. Each method is evaluated by comparing the saliency heatmaps for the predictions of network 1 and 2, where $\bx_2 ^i$ is simply the mean shifted input ($\bx^i_1 + \bm_2$). A saliency method that satisfies input invariance will produce identical saliency heatmaps for Network 1 and 2 despite the constant shift in input.

\begin{figure}
    \centering
    \includegraphics[width=0.6\textwidth]{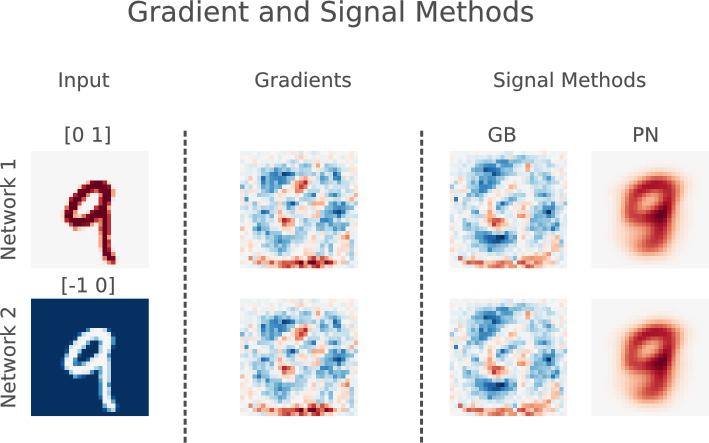}
    \caption{Evaluating the sensitivity of gradient and signal methods using MNIST with a [0,1] encoding for network $f_1$ and a [-1,0] encoding for network $f_2$. Both raw gradients and signal methods satisfy input invariance by producing identical saliency heatmaps for both networks.}

    \label{fig:gradientsignalmethods}
\end{figure}

\subsection{Gradient and Signal methods Satisfy Input Invariance} \label{subsection1}

Gradient and signal methods are not sensitive to a constant shift in inputs. In Fig.~\ref{fig:gradientsignalmethods} raw gradients, PatternNet (PN),~\citep{Kindermans2017} and GuidedBackprop (GB)~\citep{Springenberg2014} produce identical saliency heatmaps for both networks. Intuitively, gradient, PN and GB satisfy input invariance given that we are comparing two networks with an identical $f(x)$. All three methods determine attribution entirely as a function of the network/pattern weights and thus will be input invariant as long as we are comparing networks with identical weights.

In the same manner, we can say that these methods will not be input invariant when comparing networks with different weights (even if we consider models with different architectures but identical predictions for every input).

\begin{figure}
    \centering
    \includegraphics[width=\textwidth]{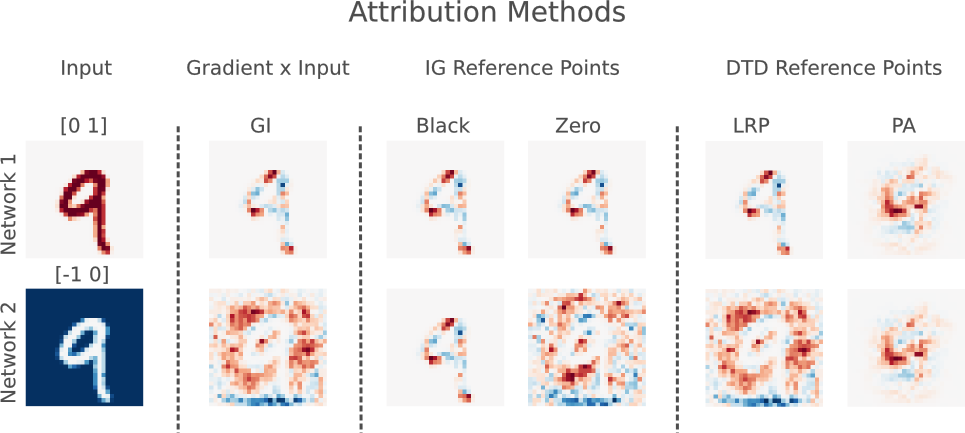}
    \caption{Evaluation of attribution method sensitivity using MNIST with a [0,1] encoding for network $f_1$ and a [-1,0] encoding for network $f_2$. Gradient x Input, IG and DTD with a zero reference point, which is equivalent to LRP \citep{Bach2015,Montavon2017}, do not satisfy input invariance and produce different attributions for each network. IG with a black image reference point and DTD with a PA reference point are not sensitive to a mean shift in input.}

    \label{fig:attributionmethods}
\end{figure}

\subsection{The Sensitivity of Attribution Methods} \label{subsection2}

We evaluate the following attribution methods: gradient times input (GI), integrated gradients (IG,~\cite{Mukund2017}) and the deep-taylor decomposition (DTD,~\cite{Montavon2017}). 

In \ref{subsection3} we find GI to be sensitive to constant shifts in the input. In \ref{subsection4} we group discussion of IG and DTD under "reference point" methods because both require that attribution is done in relation to a chosen reference. We find that satisfying input invariance depends upon the choice of reference point and the type of constant shift to the input.

\subsubsection{Gradient times input is sensitive to mean shift of inputs} \label{subsection3}

We find that the multiplication of raw gradients by the image fails to satisfy input invariance. In Fig.~\ref{fig:attributionmethods} GI produces different saliency heatmaps for both networks.

In \ref{subsection1} we determined that a saliency heatmap of gradients gradient does satisy input invariance. This breaks when the gradients are multiplied with the input image.
\[
\bs_j = \frac{\partial f(\bx)_j}{\partial\bx}\odot\bx.
\]

Multiplying by the input fails to satisfy input invariance because the input shift is carried through to final attribution. Naive multiplication by the input, as noted by \citep{Smilkov2017}, also constrains attribution without justification to inputs that are not 0.

\subsubsection{Reliability of Reference Point Methods Depends on the Choice of Reference} \label{subsection4}

Both Integrated Gradients IG,~\citep{Mukund2017} and Deep Taylor Decomposition DTD,~\citep{Montavon2017} determine the importance of inputs relative to a reference point. DTD refers to this as the root point and IG terms the reference point a baseline. The choice of reference point is not determined \textit{a priori} by the method and is instead a hyperparameter of the attribution task. 

The choice of reference point determines all subsequent attribution. In Fig.~\ref{fig:varyingrefpoint} IG and DTD show different attribution depending on the choice of reference point. We show that IG and DTD only satisfy input invariance contingent on the choice of reference point and the type of transformation considered.

\paragraph{Integrated gradients} (IG)~\citep{Mukund2017} attributes the predicted score to each input with respect to a baseline $\bx_0$. This is achieved by constructing a set of inputs interpolating between the baseline and the input.

\[
\bs_j = (\bx-\bx_0)\odot\int_{\alpha=0}^{1}\frac{\partial f(\bx_0+\alpha(\bx-\bx_0))_i}{\partial\bx}d\alpha
\]

Since this integral cannot be computed analytically, it is approximated by a finite sum ranging over $\alpha\in[0,1]$.

\[
\bs_j = (\bx-\bx_0)\odot\sum_{\alpha}\frac{\partial f (\bx_0+\alpha(\bx-\bx_0))}{\partial\bx}.
\]

We evaluate whether two possible IG reference points satisfy input invariance. Firstly, we consider an image populated uniformly with the minimum pixel from the dataset ($\bx_0=min(\bx)$) (black image) and a zero vector image. In Fig.~\ref{fig:attributionmethods}, a black image reference point produces identical attribution heatmaps whereas a zero vector reference point is not input invariant.

IG using a black image reference point is not sensitive to the constant shift in input because $\bx_0=min(\bx)$ is determined after the mean shift of the input so the difference between $\bx$ and $\bx_0$ remains the same for both networks. In network 1 this is $(\bx_1) - min(\bx_1)$ and in network 2 this is $(\bx_2 + \bm_2) - min(\bx_2 + \bm_2)$. 

IG with a zero vector reference point fails to satisfy input invariance because while the difference in network 1 is $(\bx_1 - \bx_0)$, the difference in network 2 becomes $(\bx_2 + \bm_2) - \bx_0$.

\begin{figure}
    \centering
    \includegraphics[width=0.8\textwidth]{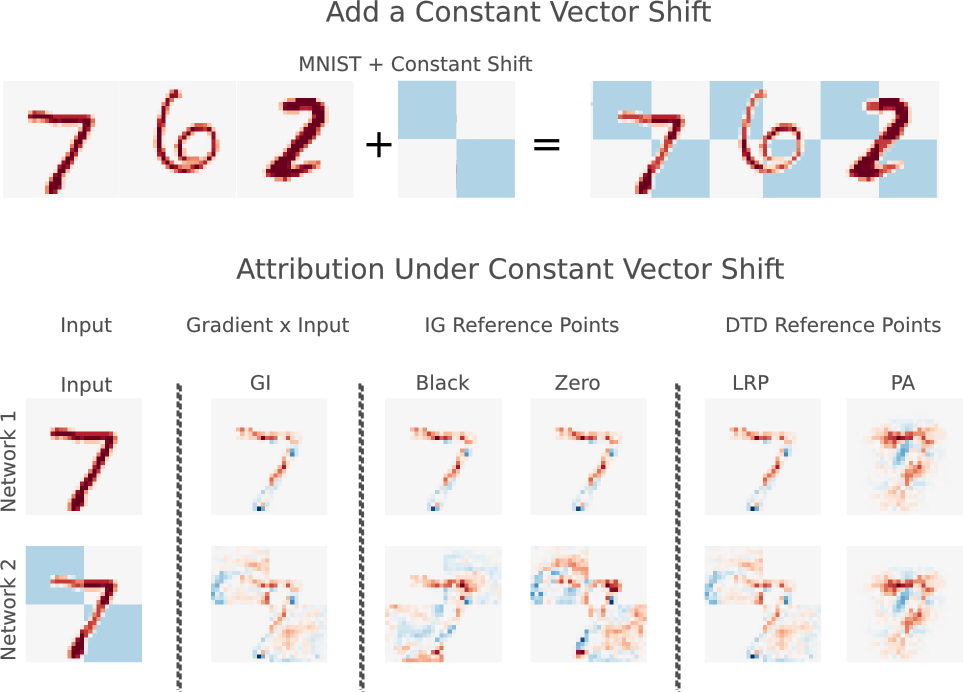}
    \caption{Evaluation of attribution method sensitivity using MNIST. Gradient x Input, all IG reference points and DTD with a LRP reference point do not satisfy input invariance and produce different attributions for each network. DTD with a PA reference point is not sensitive to the transformation of the input.}
    \label{fig:square}
\end{figure}

It is possible to construct a constant vector that will break the reliability of using a black image as a baseline.  We consider a transformation $\bx^i_2$ of the input $\bx^i_1$ where the constant vector ( $\bm_2$) added to $\bx^i_1$ is an image of a checkered box. Consistent with \textbf{Section 2} the relationship between $\bx^i_1$ and the transformed input $\bx^i_2$ is the addition of the checkered box image vector $\bm_2$.

In Fig.~\ref{fig:square} shows that we are able to manipulate the attribution heatmap of an MNIST prediction so that $\hat{\bx}$, an image of checkered boxes, appears for all reference points except for PA. This constant vector transformation causes all IG reference points to fail to satisfy input invariance.

\paragraph{Deep Taylor Decomposition (DTD)} determines attribution relative to a reference point neuron. DTD can satisfy input invariance if the right reference point is chosen. In the general formulation, the attribution of an input neuron $j$ is initialized to be equal to the output of that neuron. The attribution of other output neurons is set to zero. This attribution is backpropagated to input neurons using the following distribution rule where $s^l_j$ is the attribution assigned to neuron $j$ in layer $l$: 
 \[
s^{output}_j=y,~~~~~~~~~~~~s^{output}_{k\neq j} = 0,~~~~~~~~~~~~\bs^{l-1,j}=\frac{\bw\odot\left(\bx-\bx_0\right)}{\bw^T\bx}s^l_j.
 \]

We evaluate the input invariance of DTD using a reference point determined by Layer-wise Relevance Propagation (LRP,\cite{Bach2015}) and PatternAttribution (PA). In Fig.~\ref{fig:attributionmethods}, DTD satisfies input invariance when using a reference point defined by PA however fails to satisfy input invariance when using a reference point defined by LRP.

LRP is sensitive to the input shift because it is a case of DTD where a zero vector is chosen as the root point.\footnote{This case of DTD is called the $z-rule$ and can be shown to be equivalent to Layer-wise Relevance Propagation \citep{Bach2015,Montavon2017}. Under specific circumstances, LRP is also equivalent to the gradient times input \citep{Kindermans2016,LRPGRAD16}.}. The back-propagation rule becomes:
 
 \[
s^{output}_j=y,~~~~~~~~~~~~s^{output}_{k\neq i} = 0,~~~~~~~~~~~~\bs^{l-1,j}=\frac{\bw\odot\bx}{\bw^T\bx}s^l_j.
 \]
 
$\bs^{l-1,j}$ depends only upon the input and so attribution will change between network 1 and 2 because $\bx_1$ and$\bx_2$ differ by a constant vector.
 
PatternAttribution (PA) satisfies input invariance because the reference point $\bx_0$ is defined as the natural direction of variation in the data ~\citep{Kindermans2017}. This natural direction is determined by the covariance of the data and thus compensates explicitly for the constant vector shift of the input. Therefore it is by construction input invariant.

The PA root point is: 
\begin{equation}
\bx_0 = \bx - \ba \bw^T \bx\label{pattern_root_point}
\end{equation} where $\ba^T\bw=1$.

In a linear model:
\begin{equation}
\ba = \frac{\textrm{cov}[\bx,y]}{\bw^T\textrm{cov}[\bx,y]}.\label{linear_pattern}
\end{equation} 

For neurons followed by a ReLu non-linearity the vector $\ba$ accounts for the non-linearity and is computed as:
\[
\ba = \frac{E_+[\bx,y]-E_+[\bx]E[y]}{\bw^T(E_+[\bx,y]-E_+[\bx]E[y])}.
\]

Here $E_+$ denotes the expectation taken over values where $y$ is positive.

PA reduces to the following step:
 \[
s^{output}_i=y,~~~~~~~~~~~~s^{output}_{j\neq i} = 0,~~~~~~~~~~~~\bs^{l-1,i}=\bw\odot\ba s^l_i.
 \]
 
The vector $\ba$ depends upon covariance and thus compensates the mean shift of the input. The attribution for both networks is thus identical.

\subsection{SmoothGrad Inherits the Sensitivity Properties of Underlying Methods}

\begin{figure}
    \centering
    \includegraphics[width=0.8\textwidth]{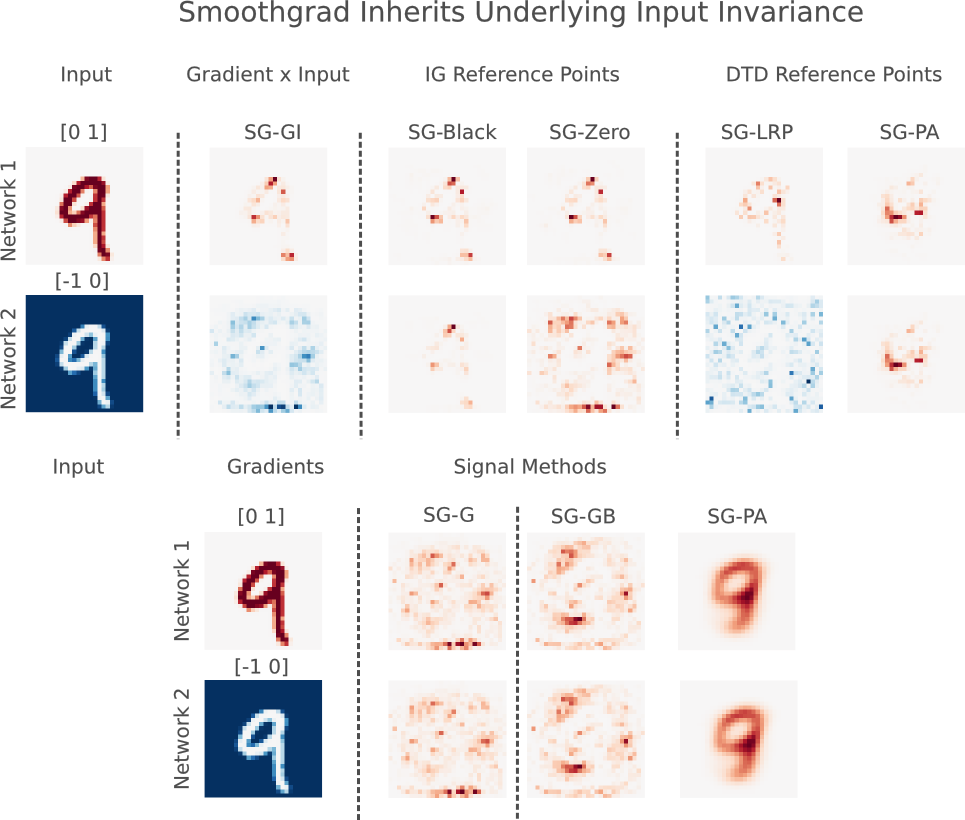}
    \caption{Smoothgrad (SG) inherits the sensitivity of the underlying attribution method. SG is not sensitive to the input transformation for gradient and signal methods (SG-PA and and SG-GB). SG does not satisfy input invariance for Integrated Gradients (SG-Zero) and Deep Taylor Decomposition (SG-LRP) when a zero vector refernce point is used. SG is invariant to the constant input shift when PatternAttribution (SG-PA) or a black image (SG-Black) are used. SG is not input invariant for gradient x input.}
    \label{fig:smoothgrad inheritence}
\end{figure}
 
 SmoothGrad (SG,~\cite{Smilkov2017}) replaces the input with ${N}$ identical versions of the input with added random noise. These noisy inputs are injected into the underlying attribution method and final attribution is the average attribution across ${N}$. For example, if the underlying methods are gradients w.r.t. the input.  $g(\bx)_j=\frac{\partial f(\bx)_j}{\partial\bx}$ SG becomes:
 
 \[
 \frac{1}{N}\sum_{i=1}^N g(\bx+\mathcal{N}(0,\sigma^2))_j
\]

SG often results in aesthetically sharper visualizations when applied to multi-layer neural networks with non-linearities. SG does not alter the attribution method itself so will always inherit the sensitivity of the underlying method to an input transformation. In Fig.~\ref{fig:smoothgrad inheritence} applying SG on top of gradients and signal methods produces identical saliency maps. SG does not satisfy input invariant when applied to gradient x input, LRP and zero vector reference points which compares SG heatmaps generated for all methods discussed so far. SG is insensitive to the input transformation when applied to PA and a black image.

\section{The Importance of Choosing an Appropriate Reference Point}
\label{subsection5}

IG and DTD satisfy input invariance when certain reference points or/and input transformations are considered. The choice of reference point is also important because it determines all subsequent attribution. In fig.\ref{fig:varyingrefpoint} attribution visually diverges for the same method if multiple reference points are considered.

A reasonable reference point choice will naturally depend upon domain and task. For example, \citep{Mukund2017} suggests that a black image is a natural reference point for image recognition tasks whereas a zero vector is a reasonable choice for text based networks. However, we have shown that the choice of reference point can lead to very different results. Unintentional misrepresentation of the model is very possible when the implications of attribution using a given reference point are unclear. Thus far, we have discussed attribution for image recognition tasks with the assumption that pre-processing steps are known and visual inspection of the points determined to be salient is possible. For audio and language based models where visual inspection is difficult or inappropriate, identifying failure points or how attribution varies under different baselines poses a challenge.

If we cannot determine the implications of reference point choice, we are limited in our ability to say anything about the reliability of the method. To demonstrate this point, we construct a constant shift of the input that takes advantage of the input invariance points of failure we have already identified.

\begin{figure}
    \centering
    \includegraphics[width=0.8\textwidth]{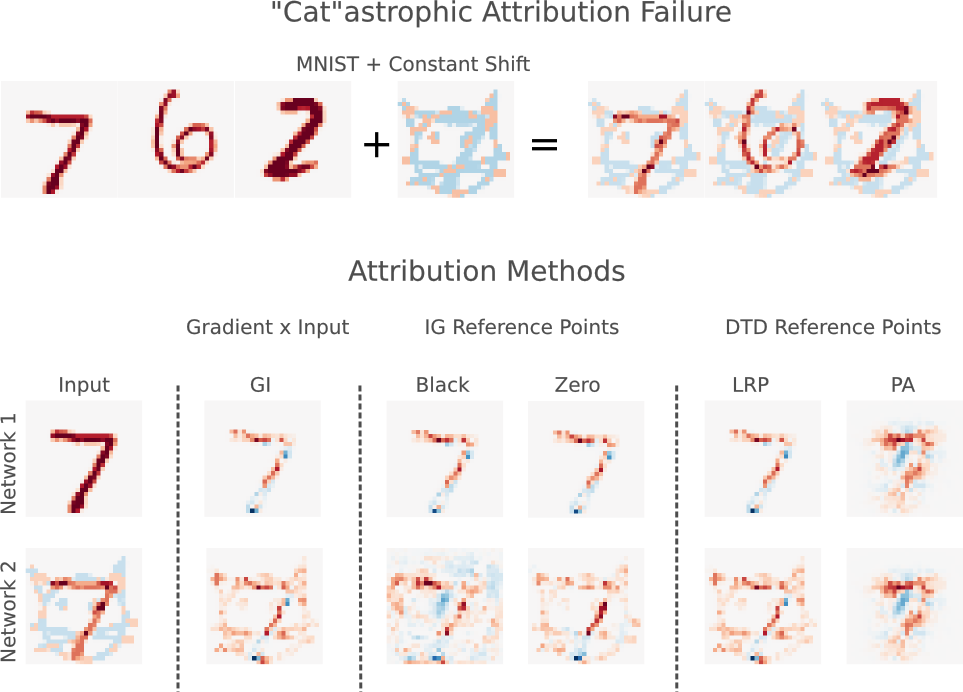}
    \caption{Evaluation of attribution method sensitivity using MNIST. Gradient x Input, IG with both a black and zero reference point and DTD with a LRP reference point, do not satisfy input invariance and produce different attribution for each network. DTD with a PA reference point is not sensitive to the transformation of the input.}
    \label{fig:kitty}
\end{figure}

In the following experiment, we construct a constant vector shift using a hand drawn image of cat. Network 1 is the same as introduced in \textbf{Section 2}. The raw image can be seen in Fig.~\ref{fig:kitty}. Consistent with \textbf{Section 2} the relationship between $\bx^i_1$ and the transformed input $\bx^i_2$ is the addition of a constant vectors $\bm_2$.

\begin{eqnarray*}
\forall i, \bx_2^i &=& \bx^i_1 + \bm_2.
\end{eqnarray*}

We construct ${\bm_2}$ by choosing a desired attribution $\hat{\bs}$ that should be assigned to a specific sample $\hat{\bx}$ when the gradient is multiplied with the input. 

${\bm_2}$ is constructed to ensure that the specific $\bx^i_2$ receives the desired attribution as follows:
\[
\bm_2 = \frac{\hat{\bs}}{\frac{\partial f_1(\bx)}{\partial \bx}}-\bx.
\]
We clip the shift to be within [-.3,.3] so that the MNIST digit is still visible, if we do not clip the end attribution would only show the cat.

In Fig.~\ref{fig:kitty} transforming the input in this manner allows purposeful misrepresentation of the attribution. All methods, except for PA, fail to satisfy input invariance and visibly show a cat as the explanation for an MNIST prediction.

How can we avoid breaks in input invariance? PA is invariant to the input transformations considered because it relies on the covariance of the data which compensates for the shift. If the data had been normalized prior to attribution, in a manner that counters this exact transformation, many of the methods considered would still satisfy input invariance. However, this is far from a systematic treatment of the reference point selection as there are input transformations outside of our experiment scope where this would not be sufficient. We believe an important research agenda is furthering the understanding of reference point choice that guarantee reliability without relying on case-by-case solutions. 

\section{Conclusion}

Saliency methods are powerful tools to gain intuition about our model. We show that numerous methods fail to attribute correctly when a constant vector shift is applied to the input. More worryingly, we show that we are able to purposefully create a deceptive explanation of the network using a hand drawn cat image. 

We introduce \textit{input invariance} as a prerequisite for reliable attribution. Our treatment of input invariance is restricted to demonstrating that there is at least one input transformation (a constant vector shift to the input) that causes numerous saliency methods to attribute incorrectly. This work is motivated by our belief that saliency methods remain valuable tools to gain intuition about the network. Understanding where they fail equips researchers with the tools to appropriately weigh the explanations these models provide. 

Guaranteeing the reliability of saliency methods is crucial in tasks where visual inspection of results is not easy or the costs of incorrect attribution is high. For example, human inspection of the attribution for an image recognition task would catch the cat attack experiment (described in section \ref{subsection5}). However, it is unclear how we would catch the same purposeful manipulation or an unintentional misrepresentation in a language or audio model where inspection is not possible or opaque. Paradoxically, these are also the cases where attribution is most needed in order to understand the data.

Determining how saliency methods fail is an important stepping stone to understanding where and how we should use these methods. An urgent research agenda, and a requirement for the use of deep neural networks in domains like medicine, is evaluating which methods and/or reference points consistently guarantee reliability for all possible transformations.

\section*{Acknowledgements}
We would like to acknowledge the thoughtful feedback and guidance of Gregoir Montavon, Mukund Sundararajan, Ankur Taly, Doug Eck and Jonas Kemp.

\bibliographystyle{iclr2018_conference}
\bibliography{iclr2018_conference.bib}
\end{document}